\documentclass{article}

\usepackage[final]{neurips_2024}

\usepackage[utf8]{inputenc} 
\usepackage[T1]{fontenc}    
\usepackage{hyperref}       
\usepackage{url}            
\usepackage{booktabs}       
\usepackage{amsfonts}       
\usepackage{nicefrac}       
\usepackage{microtype}      
\usepackage{xcolor}         

\usepackage{amsfonts,amsmath,amssymb,amsthm,mathtools}
\usepackage{natbib}
\usepackage{booktabs}

\newcommand{\bracket}[3]{\left#1 #3 \right#2}

\renewcommand{\b}{\bracket{(}{)}}
\newcommand{\sqb}{\bracket{[}{]}}

\newcommand{\bareP}{\operatorname{P}}
\newcommand{\bareQ}{\operatorname{Q}}
\renewcommand{\P}[1][]{\bareP_{#1}\b}

\newcommand{\Q}[1][]{\bareQ_{#1}\b}

\newcommand{\data}{\text{data}}

\renewcommand{\L}{\mathcal{L}}

\newcommand{\bareE}{\operatorname{E}}

\newcommand{\E}[1][]{\bareE_{#1}\sqb}

\title{Why you don't overfit, and don't need Bayes if you only train for one epoch}

\author{%
  Laurence Aitchison\\
  University of Bristol\\
  \texttt{laurence.aitchison@gmail.com} \\
}

\begin{document}

\maketitle

\begin{abstract}
Here, we show that in the data-rich setting where you only train on each datapoint once (or equivalently, you only train for one epoch), standard ``maximum likelihood'' training optimizes the true data generating process (DGP) loss, which is equivalent to the test loss.
Further, we show that the Bayesian model average optimizes the same objective, albeit while taking the expectation over uncertainty induced by finite data.
As standard maximum likelihood training in the single-epoch setting optimizes the same objective as Bayesian inference, we argue that we do not expect Bayesian inference to offer any advantages in terms of overfitting or calibration in these settings.
This explains the diminishing importance of Bayes in areas such as LLMs, which are often trained with one (or very few) epochs.
\end{abstract}

\section{Introduction}

In the early days of deep learning, datasets were small, and we trained for many epochs \citep{krizhevsky2012imagenet,he2016deep}.
This led to some degree of overfitting, as measured by poor calibration, which could be mitigated by Bayesian neural networks \citep{graves2011practical,blundell2015,gal2016dropout,sun2018functional,papamarkou2024position,unlu2020variational,ober2021global,fortuin2021bayesian} and related methods such as ensembles \citep{lakshminarayanan2017simple,fort2019deep,d2021repulsive}, dropout \citep{srivastava2014dropout,gal2016dropout,folgoc2021mc}, or methods that encourage the optimizer to find broad modes \citep{keskar2016large,dziugaite2017computing,jiang2019fantastic,foret2020sharpness,zheng2021regularizing}.
However, more recently two trends have emerged.
First, we have much larger datasets (such as multi-trillion token text datasets for LLMs e.g.\ \citealp{gao2020pile,dodge2021documenting,elazar2023s,dubey2024llama}).
With such large datasets, we often only have enough compute for a single pass over the data, or one epoch of training \citep{touvron2023llama,dubey2024llama}.
Second, overfitting in models such as LLMs appears to be far less of an issue; for instance, see Fig. 8 in \citealp{achiam2023gpt}, which shows that the pre-trained GPT-4 is well-calibrated in terms of next-token prediction probablities.
As such, in practice Bayesian or related methods such as ensembles are rarely, if ever, used in LLM pre-training, though they are sometimes used in LLM post-training \citep[e.g.]{wang2023lora,yang2023bayesian,yang2024reward}.

Why is this?
Bayesian neural networks and related methods were motivated by the need to mitigate overfitting \citep{domingos2000bayesian,cawley2007preventing,watanabe2009algebraic,izmailov2021bayesian}. 
Overfitting can be understood in terms of calibration: the tendency of neural networks to become overly certain as training proceeds through many epochs \citep{cawley2007preventing,izmailov2021bayesian}.
Here, we argue that we do not expect such overfitting, as measured by poor calibration, to occur in modern data rich training pipelines, in which data is not repeated.
In particular, we show that Bayesian inference can be understood as minimizing the expected log-likelihood under data drawn from the true data generating process, i.e.\ minimizing the expected test-loss.
Critically, we show that in the data-rich setting where we only have one epoch, we can equivalently optimize the exact same objective, simply using standard ``maximum likelihood'' training.

\section{Results}


The first question is why does standard maximum likelihood training on finite data overfit?
To make the problem concrete, consider training a neural network with weights $w$ which outputs a probability distribution, $\Q[w]{y|x}$.
For instance in classification, $\Q[w]{y|x}$ would be the distribution over labels as judged by the model, or for an LLM, $\Q[w]{y|x}$ would be the distribution over the next token.
The training loss is,
\begin{align}
  \L_\text{empirical}(w; \theta^*) &= -\E[{\P[\text{empirical}]{y|x}\P[\text{empirical}]{x}}]{\log \Q[w]{y|x}}.
  \intertext{Here, we have $\P[\text{empirical}]{y|x}$ and $\P[\text{empirical}]{x}$ because we are sampling data from a finite dataset, $\{x_i, y_i\}_{i=1}^N$.  Formally, this distribution is,}
  \P[\text{empirical}]{x} &= \frac{1}{N} \sum_{i=1}^N \delta(x-x_i),\\
  \intertext{Then taking the observed value for $y$ at that datapoint (assuming there was only one input at each location)}
  \P[\text{empirical}]{y|x} &= \delta(y-y_i).
  \intertext{Now, the optimal neural network $\Q[w^*]{y|x}$, when evaluated on the training points is,}
  \Q[w^*]{y|x{=}x_i} &= \delta(y-y_i).
\end{align}
Thus, all uncertainty vanishes on the input points, and this represents overfitting.

Bayes allows you mitigate overfitting in this setting by allowing you to work with the loss under the true data generating process,
\begin{align}
  \label{eq:ml}
  \L(w; \theta^*) &= -\E[\P{y|x, \theta^*} \P{x}]{\log \Q[w]{y|x}}.
\end{align}
Here, $\P{y|x, \theta^*} \P{x}$ is the true data generating process, with parameters, $\theta^*$. 
We call this the true data generating process (DGP) loss.
It is equivalent to the test loss, but we don't call it the test loss to avoid confusion later.
Of course, we don't know the true DGP, so we can't sample from $\P{y|x, \theta^*} \P{x}$, and hence we can't evaluate this objective.
Nonetheless, Bayes decision theory tells us how to handle this setting: we should minimize the expected loss under the posterior, $\P{\theta^*| \data}$, given finite data,
\begin{align}
  \label{eq:Eml}
  \L(w) &= \E[\P{\theta^*| \data}]{\L(w; \theta^*)}\\
  \L(w) &= -\E[\P{\theta^*| \data}]{\E[\P{y|x, \theta^*}\P{x}]{\log \Q[w]{y|x}}}
\end{align}
This expected loss can be understood as maximum likelihood on $\Q[w]{y|x}$, where we generate $y$'s by: 
\begin{align}
  x &\sim \P{x}\\
  \theta^* &\sim \P{\theta^*| \data}\\
  y &\sim \P{y|x, \theta^*}
\end{align}
Thus, the value of $\Q[w]{y|x}$ that optimizes the expected test loss is the Bayesian model average,
\begin{align}
  \bareQ^*\b{y|x} &= \int d\theta^* \P{y|x, \theta^*} \P{\theta^*| \data}.
\end{align}
Importantly, this argument isn't trying to justify Bayes using Bayes decision theory: that would be silly.
Instead, we're trying to understand the Bayesian model average as the solution to an optimization problem, namely optimizing the expected test loss.

Now this raises a question: in the data-rich, single-epoch setting where we run only one epoch, can we avoid Bayes and instead do something simpler to optimize the true DGP loss Eq.~\eqref{eq:ml}?
Remarkably, in the single epoch setting the answer is yes: the data itself is sampled from the true DGP, $\P{y| x, \theta^*} \P{x}$.
Thus, you can obtain unbiased estimates of the true DGP loss (Eq.~\ref{eq:ml}) stochastic gradient descent, on a maximum likelihood objective, with a minibatch of data of size $B$,
\begin{align}
  \label{eq:sample_estimator}
  \tilde{\L}(w) &= -\frac{1}{B} \sum_{i=1}^B \log \Q[w]{y_i|x_i}.
\end{align}
This is just the standard objective used e.g.\ in LLM pretraining.
As this estimator is unbiased, gradient descent with sufficiently small learning rates will find a (local) optimum of the true DGP objective, (Eq.~\ref{eq:ml}) which is also, remember, the objective optimized by Bayesian inference.
Thus, in the single-epoch, data-rich setting, there is no good reason to believe that Bayesian inference will give any improvements over standard maximum-likelihood pretraining in terms of overfitting or calibration.

You may be asking why we can't apply the same argument in the multi-epoch setting.
After all, the $x_i$'s and $y_i$'s in the data were ultimately generated from $\P{y|x, \theta^*} \P{x}$, whether you do one epoch or multiple epochs.
There are two alternative ways of seeing why this is the case.
First, in the multi-epoch setting, the data we are actually training on are not sampled from $\P{y| x_i, \theta^*} \P{x_i}$. 
Instead they are sampled from the empirical data distribution, $\P[\text{empirical}]{x, y}$.
These are different distributions, and this is especially evident if you think about repeated data.
In the multi-epoch setting, with distribution $\P[\text{empirical}]{x, y}$, datapoints are repeated frequently.
In contrast, if you sample from the true DGP, you usually do not expect to see repeated data.
Second, we take the weights, $w_t$ to be random variables that depend on the previous datapoints, (i.e.\ $(x_1,y_1),\dotsc,(x_{t-1},y_{t-1})$.
This makes clear that the precise notion of unbiasedness we need is,
\begin{align}
  \L(w_t) &= \E{\tilde{\L}(w_t)| w_t} = -\frac{1}{B} \sum_{i=1}^B \E{\log \Q[w_t]{y_t|x_t}| w_t}.
\end{align}
i.e.\ we need to condition on the current value of the weights, $w_t$.
This is important because it emphasises that in order to get unbiased estimates, we need samples of $(x_t, y_t)$ conditioned on $w_t$ to be drawn from the true DGP,
\begin{align}
  \P{x_t, y_t| \theta^*, w_t} &= \P{y_t, x_t| \theta^*},
\end{align}
or alternatively, we need $(x_t, y_t)$ to be independent of $w_t$, conditioned on the true DGP parameters, $\theta^*$,
\begin{align}
  x_t, y_t &\perp\!\!\!\perp w_t \;\; | \;\; \theta^*.
\end{align}
We have this in the single-epoch setting, where $x_t, y_t$ are drawn from the true data generating process, and are independent of $w_t$.
But we do not have this in the multi-epoch setting, where we may have trained on the current datapoint previously, and thus there may be dependencies between $(x_t, y_t)$ and $w_t$.

\section{Conclusions}

There are several conclusions from this line of argument.

First, when we are training for a single epoch, we are in effect training on the true DGP loss, which is equivalent to the test loss. We therefore do not expect to see overfitting. Technically, we do not expect to see training increase the degree of overfitting, though of course if you start with an overfitted model and train very little, you may still have an overfitted model. In agreement with this conclusion we indeed, do not see overfitting in practice when pre-training modern LLMs on large datasets.

Second, if the key benefits of e.g.\ Bayesian neural networks are in mitigating overfitting, and if we do not expect LLM pre-training to overfit, then we would not expect explicitly Bayesian methods (e.g.\ variational Bayes etc.) to give any benefits in LLM pretraining. Given that explicitly Bayesian methods involve some additional cost, this likely means that they should not be used in LLM pretraining.

Third, of course, these arguments do not affect the usefulness of Bayesian methods in mitigating overfitting in settings where multi-epoch training is necessary to achieve acceptable performance.
However, as the field increasingly moves towards foundation models that are trained on increasingly larger datasets, (even in post-training \citealp{dubey2024llama}), we expect the usefulness of Bayesian neural networks for the field to diminish generally.

Fourth, in this context, fast (potentially Bayes-inspired) optimizers may become even more useful \citep{aitchison2020bayesian,shen2024variational}.  In addition to the obvious benefits of a faster optimizer, they also allow you to loose less performance if you allow yourself to only go over each datapoint once.

Finally, while the importance of Bayesian neural networks may diminish in machine learning, Bayes will of course remain essential in scientific and statistical settings, where understanding our uncertainty conditioned on finite data is precisely the point.

\newpage

\bibliography{refs}
\bibliographystyle{icml2024}

\end{document}